\documentstyle[epsf,aclap]{article}

\newcommand{\comment}[1]{}
\def\topline{}
\def\botline{}
\def\TO{{\rightarrow}}
\def\LABELFIG#1#2{{\begin{tabular}{c} #1\\#2\\\end{tabular}}}

\title{\vspace{-1.2cm}Minimizing Manual Annotation Cost\\
	In Supervised Training From Corpora}

\author{Sean P. Engelson \and Ido Dagan \\
Department of Mathematics and Computer Science\\
	Bar-Ilan University\\
	52900 Ramat Gan, Israel\\
 {\tt \{engelson,dagan\}@bimacs.cs.biu.ac.il}
}

\begin{document}
\bibliographystyle{/u/segel/engelson/lib/tex/fullname}

\maketitle

\vspace{-0.5in}
\begin{abstract}
Corpus-based methods for natural language processing often use
supervised training, requiring expensive manual annotation of training
corpora.  This paper investigates methods for reducing annotation cost
by {\it sample selection}. In this approach, during training the
learning program examines many unlabeled examples and selects for
labeling (annotation) only those that are most informative at each
stage. This avoids redundantly annotating examples that contribute
little new information.  This paper extends our previous work on {\it
committee-based sample selection} for probabilistic classifiers.  We
describe a family of methods for committee-based sample selection, and
report experimental results for the task of stochastic part-of-speech
tagging.  We find that all variants achieve a significant reduction in
annotation cost, though their computational efficiency differs.  In
particular, the simplest method, which has no parameters to tune,
gives excellent results.  We also show that sample selection yields a
significant reduction in the size of the model used by the tagger.
\end{abstract}

\section{Introduction}
\label{intro}

Many corpus-based methods for natural language processing (NLP) are
based on supervised training---acquiring information from a manually
annotated corpus.  Therefore, reducing annotation cost is an important
research goal for statistical NLP.\comment{PARAGRAPH??}
The ultimate reduction in annotation cost is achieved by unsupervised
training methods, which do not require an annotated corpus at all
\cite{Kupiec92,merialdo94,Elworthy94}. 
It has been shown, however, that some supervised training prior to the
unsupervised phase is often beneficial.  Indeed, fully
unsupervised training may not be feasible for certain tasks.
\comment{***However, fully unsupervised
training may not be feasible for certain tasks. 
In other cases it has
been shown that some supervised training prior to the unsupervised
phase is beneficial.***}
This paper investigates an approach for
optimizing the supervised training (learning) phase, which reduces the
annotation effort required to achieve a desired level of accuracy of
the trained model.

In this paper, we investigate and extend the {\em committee-based
sample selection} approach to minimizing training cost
\cite{dagan-engelson95a}.
When using sample selection, a learning
program examines many unlabeled (not annotated) examples, selecting
for labeling only those that are most informative for the learner at
each stage of training
\cite{seung-et-al92,freund-et-al93,lewis-gale94,cohn-et-al94}.
This avoids redundantly annotating many examples that contribute
roughly the same information to the learner.

Our work focuses on sample selection for training probabilistic
classifiers. In statistical NLP, probabilistic classifiers are often
used to select a preferred analysis of the linguistic structure of a
text (for example, its syntactic structure \cite{Black93}, word
categories \cite{Church88}, or word senses \cite{Gale93a}).  As a
representative task for probabilistic classification in NLP, we
experiment in this paper with sample selection for the popular and
well-understood method of stochastic part-of-speech tagging using
Hidden Markov Models.

 We first review the basic approach of committee-based sample
selection and its application to part-of-speech tagging.  This basic
approach gives rise to a family of algorithms (including the original
algorithm described in \cite{dagan-engelson95a}) which we then
describe.  First, we describe the `simplest' committee-based selection
algorithm, which has no parameters to tune.  We then generalize the
selection scheme, allowing more options to adapt and tune the approach
for specific tasks.  The paper compares the performance of several
instantiations of the general scheme, including a {\em batch}
selection method similar to that of Lewis and
Gale~\shortcite{lewis-gale94}.  In particular, we found that the
simplest version of the method achieves a significant reduction in
annotation cost, comparable to that of other versions.
We also evaluate the computational
efficiency of the different variants, and the number of unlabeled
examples they consume. Finally, we study the effect of sample
selection on the size of the model acquired by the learner.

\comment{
In \cite{dagan-engelson95a} we presented initial results for a
particular variant of the committee-based approach which depends on
three parameters that were tuned experimentally. The current paper
provides an analytical review of the basic approach and includes
several new contributions. First, we describe the `simplest'
committee-based selection algorithm, which has no parameters to tune.
This algorithm is shown to achieve a comparable reduction in
annotation cost to that of the original algorithm, while being simpler
to implement and computationally more efficient. Then we generalize
the selection scheme, giving more options to adapt and tune the
approach for a specific task.  The paper compares the performance of
several instantiations of the general scheme, including the original
variant and a {\em batch} selection method similar to that of Lewis
and Gale~\shortcite{lewis-gale94}. We also evaluate the computational
efficiency of the different variants, and the number of unlabeled
examples they consume. Finally, we study the effect of sample
selection on the size of the final model acquired by the learner.
}

\section{Probabilistic Classification}
\label{prob-class}

This section presents the framework and terminology assumed for
probabilistic classification, as well as its instantiation for
stochastic bigram part-of-speech tagging.

A {\it probabilistic classifier} classifies input examples $e$ by
classes $c \in C$, where $C$ is a known set of possible
classes. Classification is based on a score function, $F_{M}(c,e)$,
which assigns a score to each possible class of an example.  The
classifier then assigns the example to the class with the highest
score.  $F_M$ is determined by a probabilistic model $M$. In many
applications, $F_M$ is the conditional probability function,
$P_M(c|e)$, specifying the probability of each class given the
example, but other score functions that correlate with the likelihood
of the class are often used.

In stochastic part-of-speech tagging, the model assumed is a Hidden
Markov Model (HMM), and input examples are sentences.  The class $c$,
to which a sentence is assigned is a sequence of the parts of speech
(tags) for the words in the sentence. The score function is typically
the joint (or conditional) probability of the sentence and the tag
sequence\footnote{This gives the Viterbi model~\cite{merialdo94},
which we use here.}. The tagger then assigns the sentence to the tag
sequence which is most probable according to the HMM.

The probabilistic model $M$, and thus the score function $F_M$, are
defined by a set of parameters, $\{\alpha_i\}$. During training, the
values of the parameters are estimated from a set of statistics, $S$,
extracted from a training set of annotated examples.  We denote a
particular model by $M=\{a_i\}$, where each $a_i$ is a specific value
for the corresponding $\alpha_i$.

In bigram part-of-speech tagging the HMM model $M$ contains three
types of parameters: {\it transition probabilities} $P(t_i\TO t_j)$
giving the probability of tag $t_j$ occuring after tag $t_i$, {\it
lexical probabilities} $P(t|w)$ giving the probability of tag $t$
labeling word $w$, and {\it tag probabilities} $P(t)$ giving the
marginal probability of a tag occurring.\footnote{This version of the
method uses Bayes' theorem
($P(w_i|t_i)\propto\frac{P(t_i|w_i)}{P(t_i)}$) ~\cite{Church88}.}  The
values of these parameters are estimated from a tagged corpus which
provides a training set of labeled examples (see
Section~\ref{sec:multinomials}).

\section{Evaluating Example Uncertainty}

A sample selection method needs to evaluate the expected usefulness,
or information gain, of learning from a given example. The methods we
investigate approach this evaluation implicitly, measuring an
example's informativeness as the uncertainty in its classification
given the current training
data~\cite{seung-et-al92,lewis-gale94,mackay92a}. The
reasoning is that if an example's classification is uncertain given
current training data then the example is likely to contain unknown
information useful for classifying similar examples in the future.

We  investigate the {\it committee-based} method, where the
learning algorithm evaluates an example by giving it to a {\it
committee} containing several variant models, all `consistent'
with the training data seen so far.  The more the committee members
agree on the classification of the example, the greater our certainty
in its classification.  This is because when the training data entails
a specific classification with high certainty, most (in a
probabilistic sense) classifiers consistent with the data will produce
that classification.

The committee-based approach was first proposed in a theoretical
context for learning binary non-probabilistic classifiers
\cite{seung-et-al92,freund-et-al93}.  In this paper, we extend 
our previous work \cite{dagan-engelson95a} where we applied the basic
idea of the committee-based approach to probabilistic classification.
Taking a Bayesian perspective, the posterior probability of a model,
$P(M|S)$, is determined given statistics $S$ from the training set
(and some prior distribution for the models). Committee members are
then generated by drawing models randomly from $P(M|S)$.  An example
is selected for labeling if the committee members largely disagree on
its classification. This procedure assumes that one can sample from
the models' posterior distribution, at least approximately.

To illustrate the generation of committee-members, consider a model
containing a single binomial parameter $\alpha$ (the probability of a
success), with estimated value $a$.  The statistics $S$ for such a
model are given by $N$, the number of trials, and $x$, the number of
successes in those trials.

Given $N$ and $x$, the `best' parameter value may be estimated by one
of several estimation methods. For example, the maximum likelihood
estimate for $\alpha$ is $a=\frac{x}{N}$, giving the model
$M=\{a\}=\{\frac{x}{N}\}$.  When generating a committee of models,
however, we are not interested in the `best' model, but rather in
sampling the distribution of models given the statistics. For our
example, we need to sample the posterior density of estimates for
$\alpha$, namely $P(\alpha=a|S)$.  Sampling this distribution yields a
set of estimates scattered around $\frac{x}{N}$ (assuming a uniform
prior), whose variance decreases as $N$ increases. In other words, the
more statistics there are for estimating the parameter, the more
similar are the parameter values used by different committee members.

For models with multiple parameters, parameter estimates for different
committee members differ more when they are based on low training
counts, and they agree more when based on high counts.  Selecting
examples on which the committee members disagree contributes
statistics to currently uncertain parameters whose uncertainty {\em
also} affects classification.

It may sometimes be difficult to sample $P(M|S)$ due to parameter
interdependence. Fortunately, models used in natural language
processing often assume independence between most model parameters.
In such cases it is possible to generate committee members by sampling
the posterior distribution for each independent group of parameters
separately.

\comment{
\subsection{Bigram part-of-speech tagging}
\label{implementation}
\label{multinomials}

We applied the committee-based method to bigram tagging, using the
method described by Dagan and Engelson~\shortcite{dagan-engelson95a}.
In brief, we break the HMM into a set of multinomials, and
sample values for each multinomial by drawing from truncated normal
distributions\footnote{The normal approximation, while easy to
implement, can be avoided.  The posterior probability for the
multinomial is given exactly by the Dirichlet distribution
\cite{johnson72} (which reduces to the Beta distribution in the
binomial case).}, using the method in
\cite[p.~214]{numerical-recipes}. 
For selection, we consider as an example each sequence of ambiguous
words, bounded by unambiguous words (those with only one possible part
of speech). (The method will be more fully described in the final
paper.) 
}

\section{Bigram Part-Of-Speech Tagging}
\label{implementation}

\subsection{Sampling model parameters}
\label{sec:multinomials}

In order to generate committee members for bigram tagging, we sample
the posterior distributions for transition probabilities, $P(t_i\TO
t_j)$, and for lexical probabilities, $P(t|w)$ (as described in
Section \ref{prob-class}).

Both types of the parameters we sample have the form of multinomial
distributions.  Each multinomial random variable corresponds to a
conditioning event and its values are given by the corresponding set
of conditioned events.  For example, a transition probability
parameter $P(t_i\TO{t_j})$ has conditioning event $t_i$ and
conditioned event $t_j$.

Let $\{u_i\}$ denote the set of possible values of a given multinomial
variable, and let $S = \{n_i\}$ denote a set of statistics extracted
from the training set for that variable, where $n_i$ is the number of
times that the value $u_i$ appears in the training set for the
variable, defining $N=\sum_i n_i$.  The parameters whose posterior
distributions we wish to estimate are $\alpha_i=P(u_i)$.

The maximum likelihood estimate for each of the multinomial's
distribution parameters, $\alpha_i$, is $\hat\alpha_i=\frac{n_i}{N}$.
In practice, this estimator is usually smoothed in some way to
compensate for data sparseness.  Such smoothing typically reduces
slightly the estimates for values with positive counts and gives small
positive estimates for values with a zero count.  For simplicity, we
describe here the approximation of $P(\alpha_i=a_i|S)$ for the
unsmoothed estimator\footnote{In the implementation we smooth the MLE
by interpolation with a uniform probability distribution, following
Merialdo~\shortcite{merialdo94}.  Approximate adaptation of
$P(\alpha_i=a_i|S)$ to the smoothed version of the estimator is
simple.}.

We approximate the posterior $P(\alpha_i=a_i|S)$ by first assuming
that the multinomial is a collection of independent binomials, each of
which corresponds to a single value $u_i$ of the multinomial; we
then normalize the values so that they sum to $1$.  For each such binomial, we
approximate $P(\alpha_i=a_i|S)$ as a truncated normal distribution
(restricted to [0,1]), with estimated mean $\mu=\frac{n_i}{N}$ and
variance $\sigma^2=\frac{\mu(1-\mu)}{N}$.
\footnote{The normal approximation,
while easy to implement, can be avoided.  The posterior probability
$P(\alpha_i=a_i|S)$ for the multinomial is given exactly by the
Dirichlet distribution \cite{johnson72} (which reduces to the Beta
distribution in the binomial case).  In this work we assumed a uniform
prior distribution for each model parameter; we have not addressed the
question of how to best choose a prior for this problem.}
                           
To generate a particular multinomial distribution, we randomly choose
values for the binomial parameters $\alpha_i$ from their approximated
posterior distributions (using the simple sampling method given in
\cite[p.~214]{numerical-recipes}), and renormalize them so that they
sum to 1.  Finally, to generate a random HMM given statistics $S$, we
choose values independently for the parameters of each multinomial,
since all the different multinomials in an HMM are independent.

\comment{
Both types of the parameters we sample have the form of multinomial
distributions.  
Since the multinomials in an HMM are independent, the
posterior distributions of their values can be sampled separately. The
posterior distribution for each multinomial are approximated by first
assuming that the multinomial is a collection of independent
binomials, each corresponding to a single value of the multinomial; we
then separately sample each binomial \cite{numerical-recipes} and
renormalize the parameters of all these binomials such that they would
sum to $1$.   

For each binomial, the posterior distribution  of its
parameter is approximated as a truncated normal distribution
(restricted to [0,1]), 
with estimated mean $\mu=\frac{x}{n}$ and variance
$\sigma^2=\frac{\mu(1-\mu)}{n}$.\footnote{The normal approximation,
while easy to implement, can be avoided.  The posterior probability
$P(\alpha_i=a_i|S)$ for the multinomial is given exactly by the
Dirichlet distribution \cite{johnson72} (which reduces to the Beta
distribution in the binomial case).}  
}

\subsection{Examples in bigram training}

Typically, concept learning problems are formulated such that there is
a set of training examples that are independent of each other.  When
training a bigram model (indeed, any HMM), this is not true,
as each word is dependent on that before it.  This problem is solved
by considering each sentence as an individual example.  More 
generally, it is possible to break the text at any point where tagging is
unambiguous.   We thus use
unambiguous words (those with only one possible part of speech) as
example boundaries in bigram tagging.  This allows us to train on
smaller examples, focusing training more on the truly informative
parts of the corpus.

\section{Selection Algorithms}

Within the committee-based paradigm there exist different methods for
selecting informative examples.  Previous research in sample selection
has used either {\it sequential} selection
\cite{seung-et-al92,freund-et-al93,dagan-engelson95a}, or {\it batch}
selection \cite{lewis-catlett94,lewis-gale94}.   We 
describe here general algorithms for both sequential and batch selection.

{\it Sequential} selection examines unlabeled examples as they are
supplied, one by one, and measures the disagreement in their
classification by the committee.
Those examples determined to be sufficiently informative are selected
for training. Most simply, we can use a committee of size two
and select an example
when the two models disagree on its classification.  This gives the
following, parameter-free, {\bf two member sequential selection
algorithm}, executed for each unlabeled input example $e$:
\begin{enumerate}\itemsep=0cm\parsep=0cm
\item Draw 2 models randomly from $P(M|S)$, where $S$ are
statistics acquired from previously labeled examples;

\item Classify $e$ by each model, giving classifications $c_1$ and
$c_2$;

\item If $c_1 \neq c_2$, select $e$ for annotation;

\item If $e$ is selected, get its correct label and update $S$
accordingly.
\end{enumerate}
This basic algorithm needs no parameters.  If desired, it is possible to
tune the frequency of selection, by changing the variance of $P(M|S)$
(or the variance of $P(\alpha_i=a_i|S)$ for each parameter),
where larger variances increase the rate of disagreement among the
committee members.  We implemented this effect by employing a temperature
parameter $t$, used as a multiplier of the variance of the posterior
parameter distribution. 

A more general algorithm results from allowing (i)~a larger number of
committee members, $k$, in order to sample $P(M|S)$ more precisely,
and (ii)~more refined example selection criteria.  This gives the
following {\bf general sequential selection algorithm}, executed for
each unlabeled input example $e$:
\begin{enumerate}\itemsep=0cm\parsep=0cm
\item Draw $k$ models $\{M_i\}$ randomly from $P(M|S)$ (possibly using
a temperature $t$);

\item Classify $e$ by each model $M_i$ giving classifications
$\{c_i\}$;

\item Measure the disagreement $D$ over $\{c_i\}$; \label{step:disagree}

\item Decide whether to select $e$ for annotation, based on the value
of $D$; \label{step:select}

\item If $e$ is selected, get its correct label and update $S$
accordingly.
\end{enumerate}
It is easy to see that two member sequential selection is a special case of
general sequential selection, where any disagreement is considered
sufficient for selection.  In order to instantiate the general
algorithm for larger committees, we need to define (i)~a measure for
disagreement (Step~\ref{step:disagree}), and (ii)~a selection
criterion (Step~\ref{step:select}).

Our approach to measuring disagreement is to use the {\it vote
entropy}, the entropy of the distribution of classifications assigned
to an example (`voted for') by the committee members. Denoting the
number of committee members assigning $c$ to $e$ by $V(c,e)$, the vote
entropy is:
\[
   D = -\frac{1}{\log k} \sum_c \frac{V(c,e)}{k}\log \frac{V(c,e)}{k}
\]
(Dividing by $\log k$ normalizes the scale for the number of committee
members.)  Vote entropy is maximized when all committee members
disagree, and is zero when they all agree.

In bigram tagging, each example consists of a sequence of several
words.  In our system, we measure $D$ separately for each word, and
use the average entropy over the word sequence as a measurement of
disagreement for the example.  We use the average entropy rather than
the entropy over the entire sequence, because the number of committee
members is small with respect to the total number of possible tag
sequences.  Note that we do not look at the entropy of the
distribution given by each single model to the possible tags
(classes), since we are only interested in the uncertainty of the
final classification (see the discussion in
Section~\ref{sec:discussion}).

We consider two alternative selection criteria (for
Step~\ref{step:select}).  The simplest is {\it thresholded selection},
in which an example is selected for annotation if its vote entropy
exceeds some threshold $\theta$.  The other alternative is {\it
randomized selection}, in which an example is selected for annotation
based on the flip of a coin biased according to the vote entropy---a
higher vote entropy entailing a higher probability of selection.  We
define the selection probability as a linear function of vote entropy:
$p = g D$, where $g$ is an {\it entropy gain} parameter.  The
selection method we used in our earlier work
\cite{dagan-engelson95a} is randomized sequential selection using this
linear selection probability model, with parameters $k$, $t$ and $g$.

An alternative to sequential selection is {\it batch
selection}. Rather than evaluating examples individually for their
informativeness a large batch of examples is examined, and the $m$
best are selected for annotation.  The {\bf batch selection algorithm},
executed for each batch $B$ of $N$ examples, is as follows:
\begin{enumerate}\itemsep=0cm\parsep=0cm
\item For each example $e$ in $B$:
\begin{enumerate}\itemsep=0cm\parsep=0cm
\item Draw $k$ models randomly from $P(M|S)$;

\item Classify $e$ by each model, giving classifications $\{c_i\}$;

\item Measure the disagreement $D_e$ for $e$ over $\{c_i\}$;
\end{enumerate}

\item Select for annotation the $m$ examples from $B$ with the highest
$D_e$;

\item Update $S$ by the statistics of the selected examples.
\end{enumerate}
This procedure is repeated sequentially for successive batches of $N$
examples, returning to the start of the corpus at the end.  If $N$ is
equal to the size of the corpus, batch selection selects the $m$
globally best examples in the corpus at each stage (as in
\cite{lewis-catlett94}).  On the other hand, as $N$ decreases, batch
selection becomes closer to sequential selection.

\section{Experimental Results}
\label{sec:results}

This section presents results of applying committee-based
sample selection to bigram part-of-speech tagging, as compared with
complete training on all examples in the corpus.  Evaluation was
performed using the University of Pennsylvania tagged corpus from the
ACL/DCI CD-ROM I.  For ease of implementation, we used a complete
(closed) lexicon which contains all the words in the
corpus.

\begin{figure}[tbh]
\centerline{
\LABELFIG{
\epsfxsize=3.0in
\epsfysize=2.4in
  \epsfbox{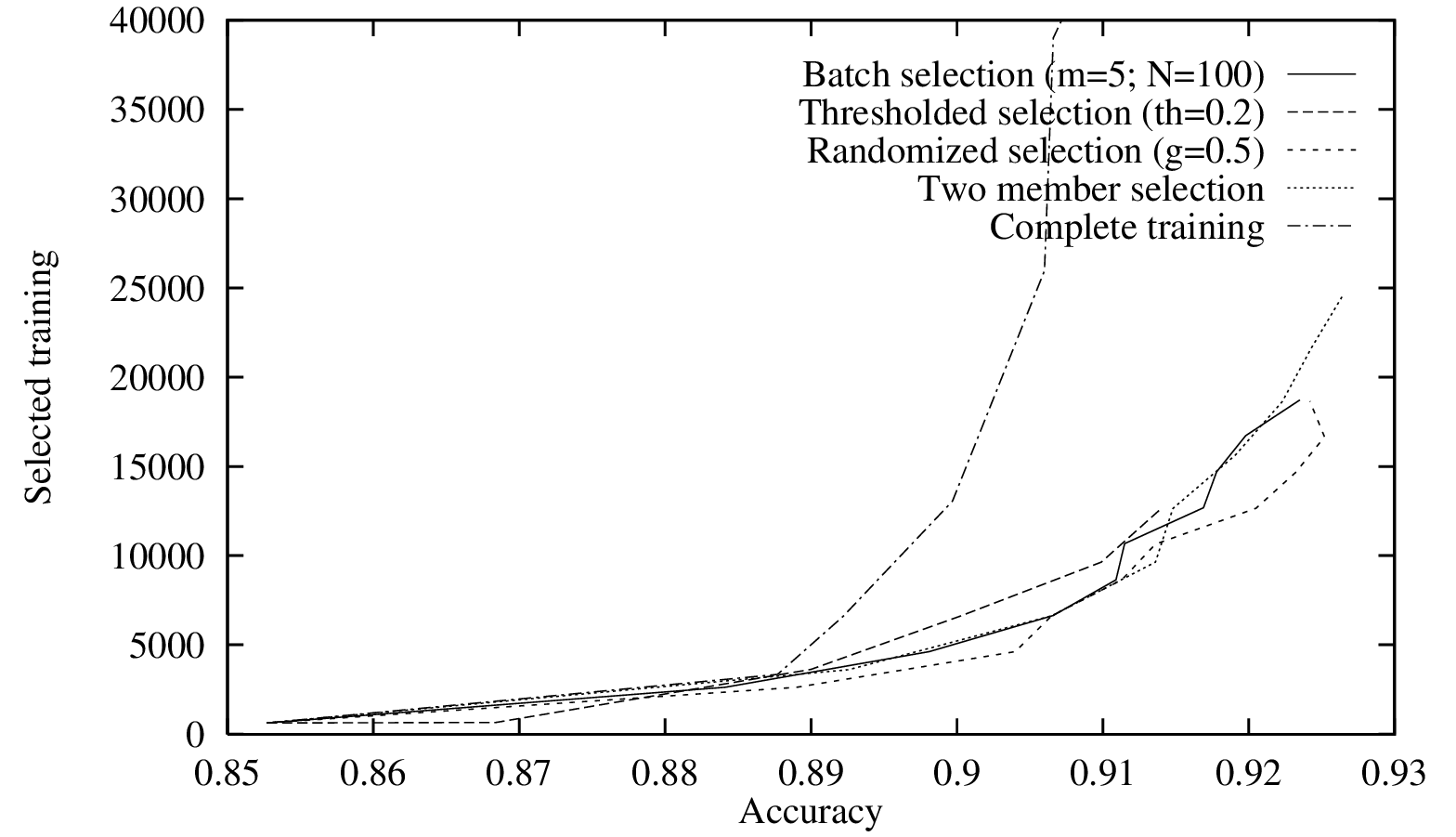}
}{(a)}
}
\centerline{
\LABELFIG{
\epsfxsize=3.0in
\epsfysize=2.3in
\epsfbox{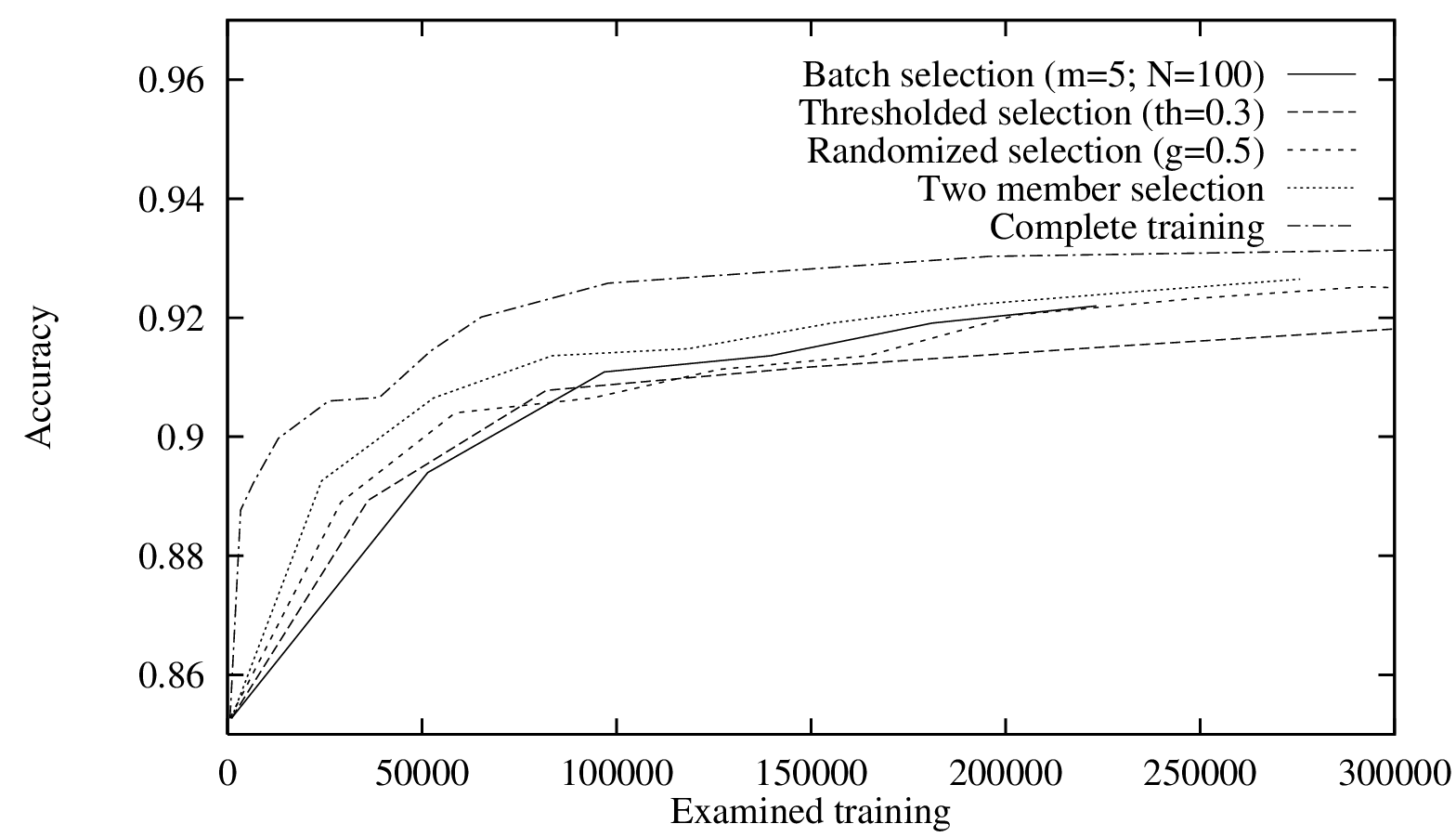}
}{(b)}
}
\caption{\footnotesize Training versus accuracy.  In batch, random, and thresholded
runs, $k=5$ and $t=50$.  (a)~Number of ambiguous words selected for
labeling versus classification accuracy achieved.  (b)~Accuracy versus
number of words examined from the corpus (both labeled and unlabeled).
\label{fig:comparison}}
\end{figure}

\begin{figure}[tbh]
\topline
\centerline{
\LABELFIG{
\epsfxsize=3.0in
\epsfysize=2.4in
  \epsfbox{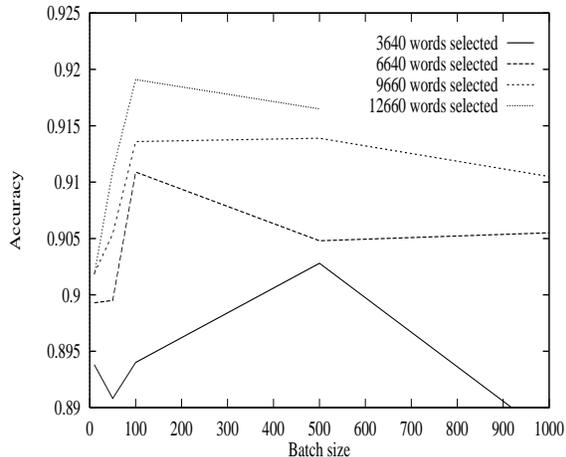}
}{(a)}
}
\centerline{
\LABELFIG{
\epsfxsize=3.0in
\epsfysize=2.3in
\epsfbox{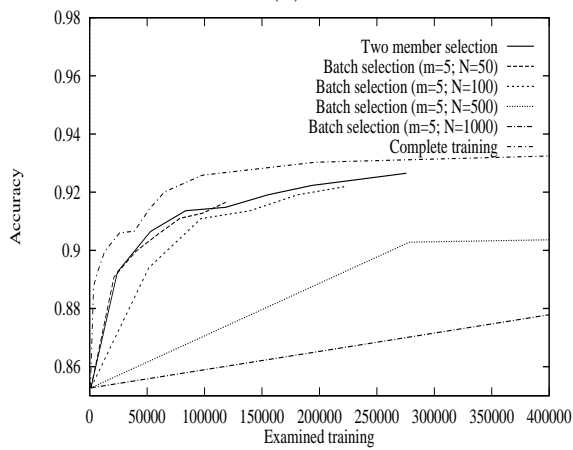}
}{(b)}
}
\caption{\footnotesize Evaluating batch selection, for $m=5$.  (a)~Accuracy achieved
versus batch size at different numbers of selected training words.
(b)~Accuracy versus number of words examined from the corpus for
different batch sizes.
\label{fig:batch}}
\botline
\end{figure}

The committee-based sampling algorithm was initialized using the first
1,000 words from the corpus, and then
sequentially examined the following examples in the corpus for
possible labeling.  The training set consisted of the first million
words in the corpus, with sentence ordering randomized to compensate
for inhomogeneity in corpus composition.  The test set was a separate
portion of the corpus, consisting of 20,000 words.  We compare the
amount of training required 
by different selection methods to achieve a given tagging accuracy on
the test set, where both the amount of training and tagging accuracy
are measured over ambiguous words.\footnote{Note that most other work on
tagging has measured accuracy over all words, not just ambiguous ones.
Complete training of our system on 1,000,000 words gave us an accuracy
of 93.5\% over ambiguous words, which corresponds to an accuracy of
95.9\% over all words in the test set, comparable to other published
results on bigram tagging.}

The effectiveness of randomized committee-based selection for
part-of-speech tagging, with 5 and 10 committee members, was
demonstrated in \cite{dagan-engelson95a}.  Here we 
present and compare results for batch, randomized, thresholded, and
two member committee-based selection.

\begin{figure}[tbh]
\topline
\centerline{
\LABELFIG{
\epsfxsize=3.0in
\epsfysize=2.4in
  \epsfbox{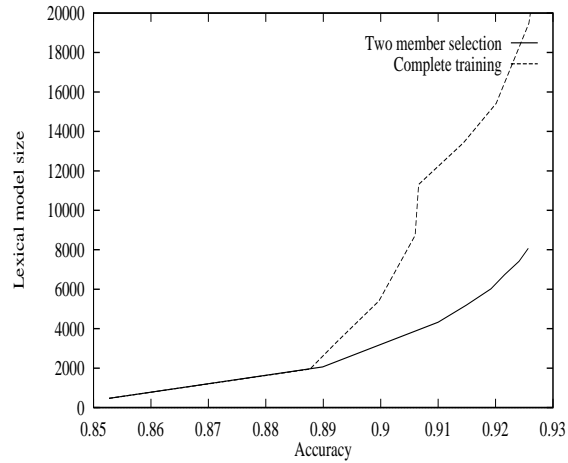}
}{(a)}
}
\centerline{
\LABELFIG{
\epsfxsize=3.0in
\epsfysize=2.3in
\epsfbox{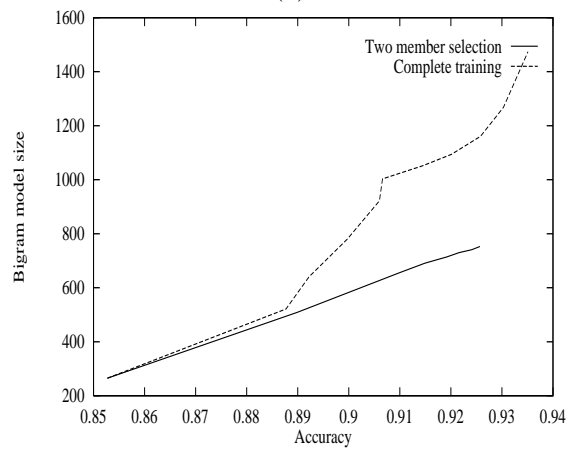}
}{(b)}
}
\caption{\footnotesize The size of the trained model, measured by the number of
frequency counts $> 0$, plotted ($y$-axis) versus classification
accuracy achieved ($x$-axis).  (a)~Lexical counts (freq($t,w$))
(b)~Bigram counts (freq($t_1\TO{t_2}$)).
\label{fig:modelsize}}
\botline
\end{figure}

Figure~\ref{fig:comparison} presents the results of comparing the
several selection methods against each other.  The plots shown are for
the best parameter settings that we found through manual tuning for
each method.  Figure~\ref{fig:comparison}(a) shows the advantage
that sample selection gives with regard to annotation cost.  For
example, complete training requires annotated examples containing
98,000 ambiguous words to achieve a 92.6\% accuracy (beyond the scale
of the graph), while the selective methods require only 18,000--25,000
ambiguous words to achieve this accuracy. We also find that, to a
first approximation, all selection methods considered give similar
results.  Thus, it seems that a refined choice of the selection method
is not crucial for achieving large reductions in annotation cost.

This equivalence of the different methods also largely holds with
respect to computational efficiency.  Figure~\ref{fig:comparison}(b)
plots classification accuracy versus number of words {\it examined},
instead of those {\it selected}.  We see that while all selective
methods are less efficient in terms of examples examined than complete
training, they are comparable to each other.  Two member
selection seems to have a clear, though small, advantage.

In Figure~\ref{fig:batch} we investigate further the properties of
batch selection.  Figure~\ref{fig:batch}(a) shows that accuracy
increases with batch size only up to a point, and then starts to
decrease.  This result is in line with theoretical difficulties with
batch selection~\cite{freund-et-al93} in that batch selection does not
account for the distribution of input examples.  Hence, once batch
size increases past a point, the input distribution has too little
influence on which examples are selected, and hence classification
accuracy decreases.  Furthermore, as batch size increases,
computational efficiency, in terms of the number of examples examined
to attain a given accuracy, decreases tremendously
(Figure~\ref{fig:batch}(b)).

The ability of committee-based selection to focus on the more
informative parts of the training corpus is analyzed in
Figure~\ref{fig:modelsize}.  Here we examined the number of lexical
and bigram counts that were stored (i.e, were non-zero) during
training, using the two member selection algorithm and complete
training.  As the graphs show, the sample selection method achieves
the same accuracy as complete training with fewer lexical and bigram
counts.  This means that many counts in the data are less useful for
correct tagging, as replacing them with smoothed estimates works just
as well.\footnote{As noted above, we smooth the MLE estimates by
interpolation with a uniform probability distribution
\cite{merialdo94}.}  Committee-based selection ignores such counts,
focusing on parameters which improve the model. This behavior has the
practical advantage of reducing the size of the model significantly
(by a factor of three here).  Also, the average count is lower in a
model constructed by selective training than in a fully trained model,
suggesting that the selection method avoids using examples which
increase the counts for already known parameters.

\section{Discussion}
\label{sec:discussion}

Why does committee-based sample selection work?  Consider the
properties of those examples that are selected for training.  In
general, a selected training example will contribute data to several
statistics, which in turn will improve the estimates of several
parameter values.  An {\it informative} example is therefore one whose
contribution to the statistics leads to a significantly useful
improvement of model parameter estimates.   Model
parameters for which acquiring additional statistics is most
beneficial can be characterized by the following three properties:
\begin{enumerate}
\item
\label{statsig}
The current estimate of the parameter is uncertain due to insufficient
statistics in the training set.  Additional statistics would bring the
estimate closer to the true value.
\item
\label{sensitive}
Classification of examples is sensitive to changes in the current
estimate of the parameter. Otherwise, even if the current value of the
parameter is very uncertain, acquiring additional statistics will not
change the resulting classifications.
\item
\label{prob}
The parameter affects classification for a
large proportion of examples in the input.  Parameters that affect only
few examples have low overall utility.
\end{enumerate}

The committee-based selection algorithms work because they tend to
select examples that affect parameters with the above three
properties.  Property~\ref{statsig} is addressed by randomly drawing
the parameter values for committee members from the posterior
distribution given the current statistics. When the statistics for a
parameter are insufficient, the variance of the posterior distribution
of the estimates is large, and hence there will be large differences
in the values of the parameter chosen for different committee members.
Note that property~\ref{statsig} is not
addressed when uncertainty in classification is only judged relative
to a {\em single} model\footnote{The use of a single model is also
criticized in \cite{cohn-et-al94}.} (as in, eg, \cite{lewis-gale94}).

Property~\ref{sensitive} is addressed by selecting examples for which
committee members highly disagree in {\it classification} (rather than
measuring disagreement in parameter estimates). Committee-based
selection thus addresses properties~\ref{statsig} and~\ref{sensitive}
simultaneously: it acquires statistics just when uncertainty in
current parameter estimates entails uncertainty regarding the {\em
appropriate} classification of the example. Our results show that this
effect is achieved even when using only two committee members to
sample the space of likely classifications.  By {\em appropriate}
classification we mean the classification given by a
perfectly-trained model, that is, one with accurate parameter values.

Note that this type of uncertainty regarding the identity of the {\em
appropriate} classification, is different than uncertainty regarding
the {\em correctness} of the classification itself.  For example,
sufficient statistics may yield an accurate $0.51$ probability
estimate for a class $c$ in a given example, making it certain that
$c$ is the appropriate classification. However, the certainty that $c$
is the {\em correct} classification is low, since there is a $0.49$
chance that $c$ is the wrong class for the example. A single model can
be used to estimate only the second type of uncertainty, which does
not correlate directly with the utility of additional training.

Finally, property~\ref{prob} is addressed by independently examining
input examples which are drawn from the input distribution.  In this
way, we implicitly model the distribution of model parameters used for
classifying input examples.  Such modeling is absent in batch
selection, and we hypothesize that this is the reason for its lower
effectiveness.

\section{Conclusions}

Annotating large textual corpora for training natural language models
is a costly process.  We propose reducing this cost significantly using 
committee-based sample selection, which reduces redundant annotation
of examples that contribute little new information. The 
method can be applied in a semi-interactive process, in which the
system selects several new examples for annotation at a time and
updates its statistics after receiving their labels from the
user. The implicit modeling of uncertainty makes the
selection system generally applicable and quite simple to implement.  

Our experimental study of variants of the selection method suggests
several practical conclusions. First, it was found that the simplest
version of the committee-based method, using a two-member committee,
yields reduction in annotation cost comparable to that of the
multi-member committee. The two-member version is simpler to
implement, has no parameters to tune and is computationally more
efficient.  Second, we generalized the selection scheme giving several
alternatives for optimizing the method for a specific task.  For
bigram tagging, comparative evaluation of the different variants of
the method showed similar large reductions in annotation cost,
suggesting the robustness of the committee-based approach.  Third,
sequential selection, which implicitly models the expected utility of
an example relative to the example distribution, worked in general
better than batch selection. The latter was found to work well only
for small batch sizes, where the method mimics sequential
selection. Increasing batch size (approaching `pure' batch selection)
reduces both accuracy and efficiency.  Finally, we studied the effect
of sample selection on the size of the trained model, showing a
significant reduction in model size.

\subsection{Further research}

Our results suggest applying committee-based sample selection to other
statistical NLP tasks which rely on estimating probabilistic
parameters from an annotated corpus.  Statistical methods for these
tasks typically assign a probability estimate, or some other
statistical score, to each alternative analysis (a word sense, a
category label, a parse tree, etc.), and then select the analysis with
the highest score. The score is usually computed as a function of the
estimates of several `atomic' parameters, often binomials or
multinomials, such as:

\begin{itemize}
\item 
In word sense disambiguation \cite{Hearst91,Gale93a}:
$P(s|f)$, where $s$ is a specific 
sense of the ambiguous word in question $w$, and $f$ is a feature of 
occurrences of $w$. Common features are words in the context of $w$ or
morphological attributes of it. 
\item
In prepositional-phrase (PP) attachment \cite{Hindle93}: $P(a|f)$,
where $a$ is a possible attachment, such as an attachment to a head verb
or noun, and $f$ is a feature, or a combination of
features, of the attachment. Common features are the words involved in
the attachment, such as the head verb or noun, the preposition, and
the head word of the PP.
\item
In statistical parsing \cite{Black93}: $P(r|h)$, the probability
of applying the rule $r$ at a certain stage of the top down derivation
of the parse tree given the history $h$ of the derivation process. 
\item
In text categorization \cite{lewis-gale94,Iwayama-tokunaga94}:
$P(t|C)$, where $t$ is a term in the document to be categorized, and
$C$ is a candidate category label.
\end{itemize}

Applying committee-based selection to supervised training for such
tasks can be done analogously to its application in the current
paper\footnote{Measuring disagreement in full syntactic parsing is
complicated. It may be approached by similar methods to those used for
parsing evaluation, which measure the disagreement between the
parser's output and the correct parse.}.  Furthermore, committee-based
selection may be attempted also for training non-probabilistic
classifiers, where explicit modeling of information gain is typically
impossible. In such contexts, committee members might be generated by
randomly varying some of the decisions made in the learning algorithm.

Another important area for future work is in developing sample
selection methods which are independent of the eventual learning
method to be applied.  This would be of considerable advantage in
developing selectively annotated corpora for general research use.
Recent work on heterogeneous uncertainty sampling
\cite{lewis-catlett94} supports this idea, using one type of model for
example selection and a different type for classification.

\paragraph{{\bf Acknowledgments.}}
We thank Yoav Freund and Yishay Mansour for helpful discussions.
The first author gratefully acknowledges the support of the Fulbright
Foundation.

\end{document}